\documentclass[10pt,letterpaper]{article}
\usepackage[top=1.2in, bottom=1.2in, left=1.5in, right=1.5in]{geometry}
\usepackage{setspace}
\usepackage{todonotes}
\usepackage{epstopdf} 
\usepackage{tabularx}
\usepackage{graphicx}
\usepackage{enumitem}
\usepackage{subfigure}
\usepackage{times}
\usepackage{url}
\setlist{nolistsep}
\usepackage{amsmath}
\usepackage{indentfirst}
\usepackage{amssymb}
\usepackage{amsfonts}
\usepackage{amsthm}
\usepackage{subfigure}
\usepackage{color}
\usepackage[nospace,noadjust]{cite}
\usepackage{verbatim}
\usepackage[]{graphicx}
\usepackage{bm}
\usepackage{hyperref}
\usepackage{algorithmic}
\usepackage{algorithm}
\newtheorem{theorem}{Theorem}

\newtheorem{corollary}{Corollary}
\newtheorem{definition}{Definition}
\usepackage{authblk}
\usepackage[automark,headsepline,plainheadsepline]{scrpage2}
\newtheorem{Lemma}{Lemma}
\newtheorem{lemma}{Lemma}
\usepackage{authblk}
\ihead{On Learning from Label Proportions}

\begin{document}
\title{On Learning from Label Proportions}

\author[1]{Felix X. Yu\thanks{yuxinnan@ee.columbia.edu}}
\author[3]{Krzysztof Choromanski}
\author[3]{Sanjiv Kumar}
\author[2]{\\Tony Jebara}
\author[1]{Shih-Fu Chang}
\affil[1]{Department of Electrical Engineering, Columbia University}
\affil[2]{Department of Computer Science, Columbia University}
\affil[3]{Google Inc.}

\date{}
\maketitle

\begin{abstract}
\thispagestyle{empty}
\emph{Learning from Label Proportions} (LLP) is a learning setting, where the training data is provided in groups, or ``bags'', and only the proportion of each class in each bag is known. The task is to learn a model to predict the class labels of the individual instances. 
LLP has broad applications in political science, marketing, healthcare, and computer vision. 
This work answers the fundamental question, \emph{when and why LLP is possible}, by introducing a general framework, 
\emph{Empirical Proportion Risk Minimization} (EPRM).  
EPRM learns an instance label classifier to match the given label proportions on the training data. 
Our result is based on a two-step analysis. First, we provide a VC bound on the generalization error of the bag proportions. We show that the bag sample complexity is only mildly sensitive to the bag size. Second, we show that under some mild assumptions, good bag proportion prediction guarantees good instance label prediction. 
The results together provide a formal guarantee that the individual labels can indeed be learned in the LLP setting. 
We discuss applications of the analysis, including justification of LLP algorithms, learning with population proportions, and a paradigm for learning algorithms with privacy guarantees. We also demonstrate the feasibility of LLP based on a case study in real-world setting: predicting income based on census data. 
\end{abstract}

\section{Introduction}

A lot of information of individuals is released in the form of group label proportions. For example, 
after election, the proportions of votes of each demographic 
area are released by the government. 
In healthcare, the proportions of diagnosed diseases 
of each ZIP code area are available to public.
Is it possible to learn a model to predict the individual labels based on only the group-level label proportions?
Recent works in a machine learning setting called \emph{Learning from Label Proportions} (LLP) tried to study this problem. In LLP, the training instances are provided as groups, or ``bags''. In this work, we consider a binary learning setting: for each bag, only the proportion of the positive instances is available. The task is to learn a model to predict the labels of individual instances. 
It is shown that by combining the label proportions 
with instance-level attributes, models capable 
of correctly predicting instance labels can be learned \cite{quadrianto2009estimating, rueping2010svm, Yu13_3}. 
As non-confidential attributes of individuals can be easily acquired from census survey, digital footprint, shopping history \emph{etc}., LLP leads to not only promising new applications, but also serious privacy concerns as releasing label proportions may result in the discovery of sensitive personal information. Recently, LLP has also been applied in visual attribute modeling \cite{mm14_sentiment, mm14_attribute} and event detection \cite{cvpr14_video} in computer vision. 

This work studies when and why individual labels can be learned from label proportions, by analyzing a general framework, namely \emph{Empirical Proportion Risk Minimization} (EPRM). EPRM optimizes the instance-level classifier to minimize the empirical proportion loss. In other words, it tries to find an instance hypothesis to match the given label proportions. 
The main contribution of this paper is a formal guarantee that under some mild assumptions, the individual instance labels can be recovered (learned), with the EPRM framework. Specifically, we provide a two-step analysis.

Our first result bounds the generalization error of bag proportions by the empirical bag proportion error (Section \ref{sec:iidbag}). We show that the sample complexity is only mildly sensitive to the bag size. In other words, our analysis shows that given enough training bags, it is possible to learn a bag proportion predictor, which generalizes well to unseen bags. 
This conclusion in itself is interesting as in some applications we are simply interested in getting good proportion estimates for bags: doctors may want to predict the rate of disease on certain geographical area, and companies may want to predict attrition rate of certain department.

Second, we show that under some mild conditions, the instance label error can be controlled by the bag proportion error (Section \ref{sec:instance}). In other words, ``good'' bag proportion predictions imply ``good'' instance label predictions. 
This finding is more crucial. For example, from privacy point of view, the ability to learn a good instance label predictor given label proportions can be of concern. 

In addition, we discuss using LLP to increase the privacy level of algorithms that aim to construct private preserving machine learning structures. We also discuss applying the analysis into justifying LLP algorithms, and LLP in some real-world cases (Section \ref{sec:applications}). Finally, we demonstrate the feasibility of LLP in a case study: predicting income based on census data (Section \ref{sec:exp}).

\vspace{-0.1cm}
\section{Related Works}
\label{sec:related}
\vspace{-0.1cm}

\textbf{Algorithms for LLP.}  
In their seminal work, \cite{quadrianto2009estimating} proposed to estimate the mean of each class using the mean of each bag and the label proportions. These estimates are then used in a
conditional exponential model to maximize the log likelihood. The algorithm is under a restrictive assumption that the instances are conditionally independent given the label. \cite{rueping2010svm}
proposed to use a large-margin regression method by assuming the mean instance of each bag having a soft label corresponding to the label proportion. 
As an extension to multiple-instance learning, \cite{Kuck05} 
designed a hierarchical probabilistic model to generate consistent label proportions. 
Similar ideas have also been 
shown in \cite{chen2006learning} and \cite{musicant2007supervised}.
A recently proposed $\propto$SVM algorithm \cite{Yu13_3} explicitly models
the latent unknown instance labels together with the known group label
proportions in a large-margin framework.
Different from the above works, this paper provides theoretical results addressing when and why bag proportion and instance labels can be learned. Our result is independent of the algorithms.

\textbf{Multiple Instance Learning.} 
A related, yet more extensively studied learning setting is \emph{Multiple Instance Learning} (MIL)
 \cite{dietterich1997solving}.
In MIL, the learner has access to bags, with their labels generated by the Boolean OR operator on the unobserved instance labels, \emph{i.e.}, a bag is positive \emph{iff} it has at least one positive instance.
The task is to learn a binary predictor for the \emph{bags}.
It has been shown that if all the instances are drawn \emph{iid} from a single distribution, MIL is as easy as learning from \emph{iid} instances with one-sided label noise \cite{blum1998note}. 
In real-world applications, the instances inside each bag can have arbitrary dependencies, or even a manifold structure. 
The learnability and sample complexity results in the above scenarios are given by \cite{sabato2012multi, sabato2012partial}, and \cite{babenko2011multiple}, respectively. 
In this paper, we use the tools in \cite{sabato2012partial} 
to analyze the generalization error of bag proportions. 
More importantly, we show that under some conditions, a good bag proportion predictor 
implies a good instance label predictor.

\section{The Learning Setting: Learning from Label Proportions}
\label{sec:setting}

Denote by $\mathcal{X}$ the domain of instance attributes, and denote by $\mathcal{Y} = \{-1, 1\}$ the domain of the instance labels.
Each bag, consisting of the instance attributes $\tilde{x} = (x_1, \cdots, x_r)$ and their corresponding labels $\tilde{y} = (y_1, \cdots, y_r)$, is generated \emph{iid} by a probability distribution $D$ over \emph{bags} $(\mathcal{X} \times \mathcal{Y})^r$.\footnote{This is a very mild assumption. Note that \emph{iid} bags do not imply \emph{iid} instances across all bags. The instances inside each bag can have arbitrary dependencies.} 
Here, $r$ is the bag size. 
For simplicity, we assume that the bags are of the same size. Our result can be easily generalized to bags with variable sizes as described in Section \ref{sec:bag_size}.

The learner does not have access to the instance labels. Instead it receives $(\tilde{x}, f(\tilde{y}))$, where, $f:  \mathcal{Y}^r  \rightarrow \mathbb{R}$, is the proportion generation function. 
In learning with label proportions, 
\begin{equation}
f(\tilde{y}) = \frac{1}{r} \sum_{i=1}^r \frac{y_i+1}{2}. 
\label{eq:f}
\end{equation}
Let the training set received by the learner be a set of $m$ bags with their proportions $S= \{(\tilde{x}_k, f(\tilde{y}_k))\}_{k=1}^m$, in which $\tilde{x}_k$ and $\tilde{y}_k$ are the instances and the (unobserved) labels for the $k$-th bag, respectively.
We use $\mathcal{H} \subseteq \mathcal{Y}^\mathcal{X}$ to denote a hypothesis class on the instances. The learning task is to find an $h \in \mathcal{H}$ which gives low prediction error for \emph{unseen instances} generated by the above process.
In the conventional supervised learning, where all the labels of the training instances are known, a popular framework is the \emph{Empirical Risk Minimization} (ERM), \emph{i.e.} finding the instance hypothesis $h \in \mathcal{H}$ to minimize the empirical instance label error. In LLP, however, the instance labels are not available at the training time. 
Therefore, we can only try to find $h \in \mathcal{H}$ to minimize the empirical proportion error.

\vspace{-0.2cm}
\section{The Framework: Empirical Proportion Risk Minimization}
\vspace{-0.1cm}

\begin{definition}
For $h \in \mathcal{H}$, define an operator to predict bag proportion based on the instances $\phi_r^f(h): \mathcal{X}^r \rightarrow \mathbb{R}$, $\phi_r^f(h)(\tilde{x}) := f\big(\big(h(x_1), \cdots, h(x_r)\big)\big), \tilde{x} \in X^r$, $\tilde{x} = (x_1, \cdots, x_r)$. And therefore the hypothesis class on the bags $\phi_r^f(\mathcal{H}) := \{\phi_r^f(h)|h \in \mathcal{H}\}$.
\end{definition}

The \emph{Empirical Proportion Risk Minimization} (EPRM) selects the instance label hypothesis $h \in \mathcal{H}$ to minimize the empirical bag proportion loss on the training set $S$. It can be expressed as follows.
\begin{align}
\arg\min_{h \in \mathcal{H}} \sum_{(\tilde{x}, f(\tilde{y})) \in S} L \left( \phi_r^f(h)(\tilde{x}), f(\tilde{y}) \right)
\end{align}
Here, $L$ is a loss function to compute the error of the predicted proportion $\phi_r^f(h)(\tilde{x})$, and the given proportion $f(\tilde{y})$.  In this paper, we assume that $L$ is 1-Lipschitz in 
$\xi: = \phi_r^f(h)(\tilde{x}) - f(\tilde{y})$. 
EPRM is a very general framework for LLP. One immediate question is whether the instance labels can be learned by EPRM. In the following sections, we provide affirmative results. We first bound the generalization error of bag proportions based on the empirical bag proportions. 
We show that the sample complexity of learning bag proportions is only mildly sensitive to the bag size (Section \ref{sec:iidbag}). 
We then show that, under some mild conditions, instance hypothesis $h$ which can achieve low error of bag proportions with high probability, is guaranteed to achieve low error on instance labels with high probability (Section \ref{sec:instance}). 

\vspace{-0.1cm}
\section{Generalization Error of Predicting the Bag Proportions}
\label{sec:iidbag}
\vspace{-0.1cm}
Given a training set $S$ and a hypothesis $h \in \mathcal{H}$, denote by $er_S^L(h)$ the empirical bag proportion error with a loss function $L$, and denote by $er_D^L(h)$ the generalization error of bag proportions with a loss function $L$ over distribution $D$:
\begin{equation*}
er_S^L(h) = \frac{1}{|S|} \sum_{(\tilde{x}, f(\tilde{y})) \in S} L(\phi_r^f(h)(\tilde{x}), f(\tilde{y})),\quad er_D^L(h) = \mathbb{E}_{(\tilde{x}, \tilde{y}) \sim D} L(\phi_r^f(h)(\tilde{x}), f(\tilde{y})).
\end{equation*}
In this section, we show that good proportion prediction is possible for unseen bags. 
Note that learning the bag proportion is basically a regression problem on the bags. Therefore, without considering the instance label hypothesis, for a smooth loss function $L$, the generalization error of bag proportions can be bounded in terms of the empirical proportion error and some complexity measure, \emph{e.g.}, fat shattering dimension \cite{anthony2009neural}, of the hypothesis class on the bags. 
Unfortunately, the above does not provide us insights into LLP, as it does not utilize the structure of the problem. As we show later, such structure is important in relating the error of bag proportion to the error of instance labels.

Based on the definitions in Section \ref{sec:setting}, our main intuition is that the complexity (or ``capacity'') of bag proportion hypothesis class $\phi_r^f(\mathcal{H})$ should be dependent on the complexity of the instance label hypothesis class $\mathcal{H}$. 
Formally, we adapt the MIL analysis in \cite{sabato2012multi, sabato2012partial}, to bound the \emph{covering number} \cite{anthony2009neural} of $\phi_r^f(\mathcal{H})$, by the covering number of $\mathcal{H}$. 
As in our case $\mathcal{H}$ is a binary hypothesis class, we further bound the covering number of $\mathcal{H}$ based on its \emph{VC-dimension} \cite{vapnik1971uniform}. 
This leads to the following theorem on the generalization error of learning bag proportions.

\begin{theorem}
\label{thm:bag}
For any $0< \delta < 1$, $0<\epsilon <1$, $h \in \mathcal{H}$, with probability at least $1-\delta$, $\text{er}_D^L(h) \leq \text{er}_S^L(h) + 
\epsilon$, if 
\vspace{-0.2cm}
\begin{equation}
m \geq \frac{64}{\epsilon^2}(2VC(\mathcal{H})\ln(12r/\epsilon) + \ln(4/\delta)),
\end{equation}
in which $VC(\mathcal{H})$ is the VC dimension of the instance label hypothesis class $\mathcal{H}$, and $m$ is the number of training bags, and $r$ is the bag size.
\end{theorem}

The details of the proof are given in the appendix. 
From the above, the generalization error of bag proportions can be bounded by the empirical proportion error if there are sufficient number of bags in training.
Note that the sample complexity (smallest sufficient size of $m$ above) grows in terms of the bag size. This is intuitive as larger bags create more ``confusions''. Fortunately, the sample complexity grows at most logarithmically with $r$. It means that the generalization error is only mildly sensitive to $r$. 
We also note that the above theorem generalizes to the well-known result of binary supervised learning, as shown below.

\begin{corollary}
\label{corollary:bag}
When bag size $r=1$, for any $0< \delta < 1$, $0<\epsilon <1$, $h \in \mathcal{H}$, with probability at least $1-\delta$, $\text{er}_D^L(h) \leq \text{er}_S^L(h) + 
\epsilon$, if 
\vspace{-0.2cm}
\begin{equation}
m \geq \frac{64}{\epsilon^2}(2VC(\mathcal{H})\ln(12/\epsilon) + \ln(4/\delta)).
\end{equation}
\end{corollary}

\section{Bounding the Instance Label Error by Bag Proportion Error}
\label{sec:instance}

From the analysis above, we know that
the generalization error of bag proportions can be bounded. The result is without any additional assumptions other than that the bags are \emph{iid}.
In this section, based on assumptions about the proportions (Section \ref{subsec:purity}), or the instances (Section \ref{subsec:conditionaliid}, Section \ref{subsec:kappa}), we present results bounding the generalization error of predicting instance labels by the generalization error of predicting bag proportions. We also discuss the limitations of LLP under our framework, providing insights into protecting the instance labels when releasing label proportions. 

The analysis in this section is based on the assumption that we already have an instance hypothesis $h$, which predicts the proportions well: $\mathbb{P} \left( | \phi_r^f(h)(\tilde{x}) -  f(\tilde{y}) | \leq \epsilon \right) \geq 1-\delta$ with some small $0< \epsilon, \delta < 1$. 
From Section \ref{sec:iidbag}, the above is true when we have sufficient number of training bags, and a good algorithm to achieve small empirical bag proportion error. 
Formally, from Theorem \ref{thm:bag}, suppose we have a hypothesis $h$, such that $er_D^L(h) \leq \epsilon'$. 
Then $\mathbb{P}_{(\tilde{x}, \tilde{y}) \sim D}(|\phi_r^f(h)(\tilde{x}) - f(\tilde{y}) | \leq \epsilon'' )$ for some other small $\epsilon''$ can also be bounded. With Markov's inequality:  for any $0 < \delta < 1$, define $\epsilon'' = \epsilon'/ \delta$, then $\mathbb{P}_{(\tilde{x},\tilde{y}) \sim D}(| \phi_r^f(h)(\tilde{x}) - f(\tilde{y}) | \leq  \epsilon'') \geq 1-\delta$.

\subsection{Learning with Pure Bags}
\label{subsec:purity}

Intuitively, if the bags are very ``pure'', \emph{i.e.}, their proportions are either very low or very high, the instance label error can be well controlled by the bag proportion error. The case that all the bags are with proportions 0 or 1 should be the easiest, as it is identical to the conventional supervised learning setting. We justify the intuition in this section. 
For $0 < \eta < 1$, we say that a bag is \emph{$(1-\eta)$-pure} if at least a fraction $(1-\eta)$ of all instances have the same label. The following theorem summarizes our analysis.
\begin{theorem}
\label{thm:instance}
Let $h$ be a hypothesis satisfying $\mathbb{P}_{(\tilde{x},\tilde{y}) \sim D}(| \phi_r^f(h)(\tilde{x}) - f(\tilde{y}) | \leq  \epsilon) \geq 1-\delta$ for some $0 < \epsilon, \delta < 1$. Assume  that the probability that
each bag is \emph{$(1-\eta)$-pure} is at least $1-\rho$ for some $0 < \eta, \rho < 1$. 
Then, for any $0 < \tau < 1$, the probability that $h$ classifies correctly at least a fraction $(1-\tau)(1-\delta-\rho)(1-2\eta-\epsilon)$ of all $nr$ instances given in $n$ bags is at least $1-e^{-\frac{\tau^2}{2} nr(1-\delta-\rho)(1-2\eta-\epsilon)}$. 
\end{theorem}
Theorem \ref{thm:instance} follows from Lemma \ref{lemma:instance} with standard concentration inequalities (the proof is given in the appendix). In addition, Theorem \ref{thm:instance} is shown to be tight based on Lemma \ref{lemma:tight}.
\begin{lemma}
\label{lemma:instance}
Let $h$ be a hypothesis satisfying 
$\mathbb{P}_{(\tilde{x},\tilde{y}) \sim D}(| \phi_r^f(h)(\tilde{x}) - f(\tilde{y}) | \leq  \epsilon) \geq 1-\delta$ 
for some $0 < \epsilon, \delta < 1$. Assume  that the probability that
a bag is \emph{$(1-\eta)$-pure} is at least $1-\rho$ for some $0 < \eta, \rho < 1$.
Then for that bag, the probability that $h$ classifies correctly at least a fraction $(1-2\eta-\epsilon)$ of its instances is at least
$(1-\delta-\rho)$.
\end{lemma}

\begin{lemma}
\label{lemma:tight}
There exists a distribution $\mathcal{D}$ over all bags of size $r$ and a learner $h$ such that
$\phi_r^f(h)(\tilde{x}) = f(\tilde{y})$, each bag is \emph{$(1-\eta)$-pure} but $h$ misclassifies a fraction $2\eta$ instances of each bag. 
\end{lemma}

Lemma \ref{lemma:tight} also shows limitations of LLP. It is interesting to consider an extreme case, where all the bags are with label proportion 50\% (they are the least ``pure''). Then there exists a hypothesis $h$ which can achieve zero bag proportion prediction error, yet with 100\% instance label error. In other words, it is hopeless to learn or recover the instance labels. This provides an interesting failure case of LLP. In the appendix, we further show a case under generative assumptions, where EPRM fails when all the bag proportions are the same and close to 50\%. Such negative results worth further study as they provide insights into ``protecting'' the instance labels when releasing label proportions. 

\subsection{Instances Are Conditionally Independent Given Bag}
\label{subsec:conditionaliid}

The purity result presented in the former section is without any assumptions on the way the bags are formed. 
In this section, we consider the case where the instances are conditionally independent given the bag.
A lot of real-world applications follow this assumption. For example, to model the voting behavior, each bag is generated by randomly sampling a number of individuals from certain location. It is reasonable to assume that the individuals are \emph{iid} given the location. 

Formally, we assume that the bags are generated from $\mathcal{D}$, a distribution over bags.
$\mathcal{D}$ is a ``mixture'' of multiple components $D_1, ..., D_M$, 
where each $D_i$ is also a distribution over bags.  
We consider the process of drawing a bag from 
$\mathcal{D}$ as firstly picking a distribution 
$D_i$ with some fixed probability $\zeta_i$, 
and then generating a bag from $D_i$.
We assume for $D_i$, $i = 1,\cdots, M$, there exists an instance distribution $D'_i$, such that generating a bag from $D_i$ is by drawing $r$ \emph{iid} instances from $D'_i$. 
Also assume that the priors $\mathbb{P}_{(x,y) \sim D'_i}(y_i=1) = \alpha_{i}$.
We note first that under the above assumptions, the purity results in the former section can be utilized. For $c > 0$, the probability that a bag generated from the mixture of multiple components 
is \emph{$(1-\eta)$-pure}, is at least  $1-e^{-2rc^2}$, where $\eta = \max_{i} (\min(\alpha_{i},1-\alpha_{i})) + c$. The proof is straightforward with Chernoff bound. One disadvantage of the purity result is that it is only useful when $\alpha_i, \forall i$ is not close to 1/2.

\begin{figure*}[!t]
\centering
\small
\subfigure[$\epsilon = 0$]
{\includegraphics[width = 3.4cm]{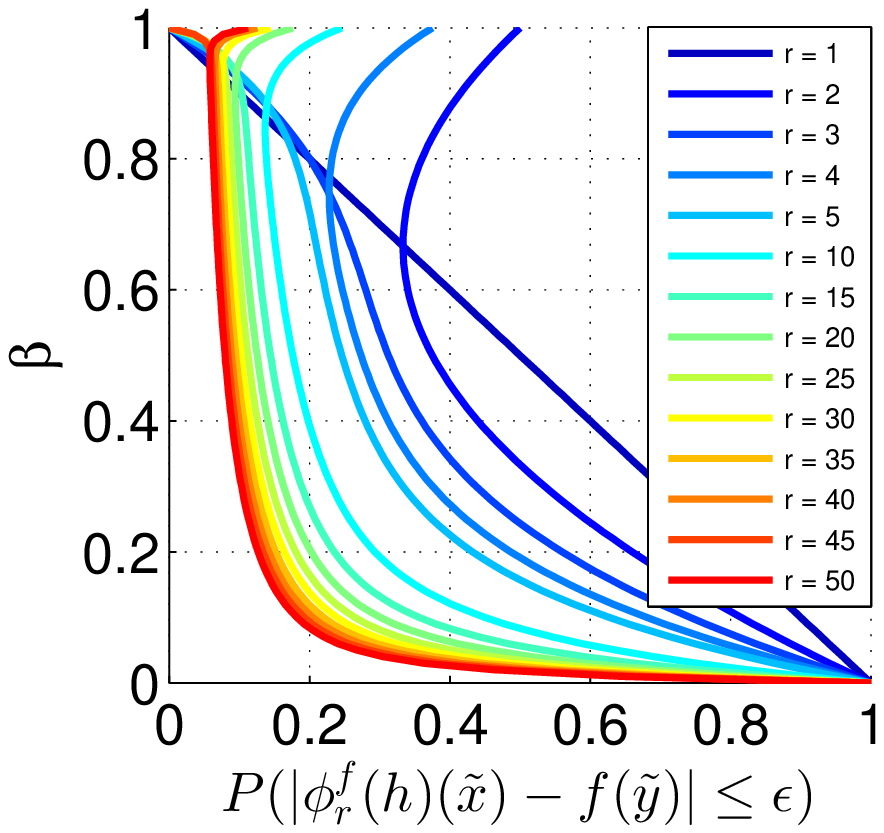}}
\subfigure[$\epsilon = 0.1$]
{\includegraphics[width = 3.4cm]{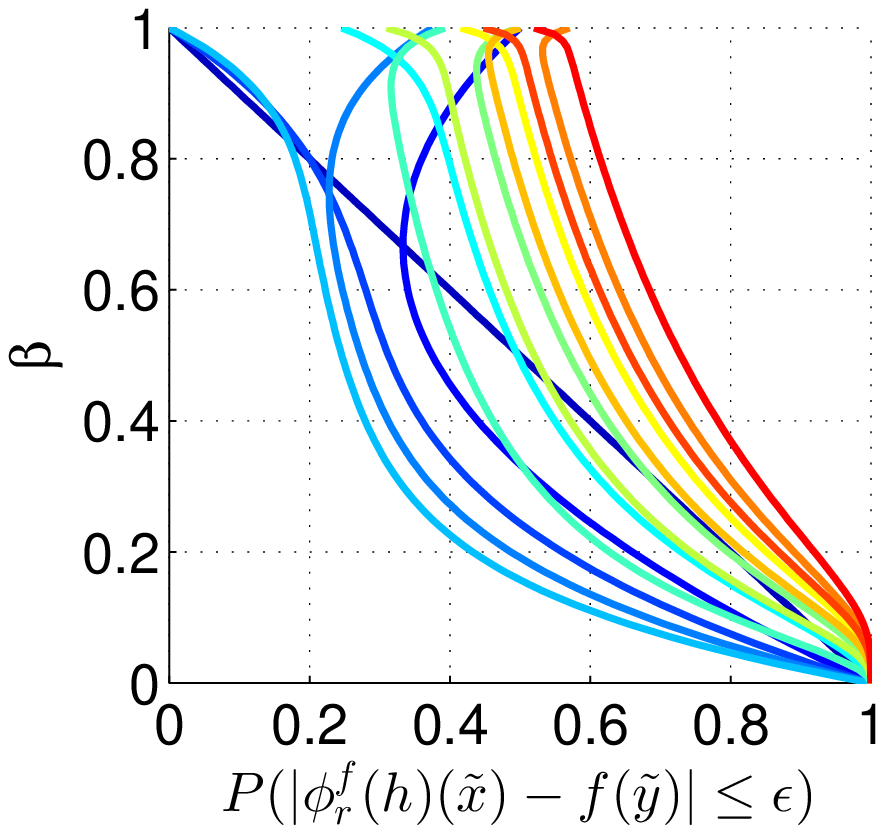}}
\subfigure[$r = 50$]
{\includegraphics[width = 3.4cm]{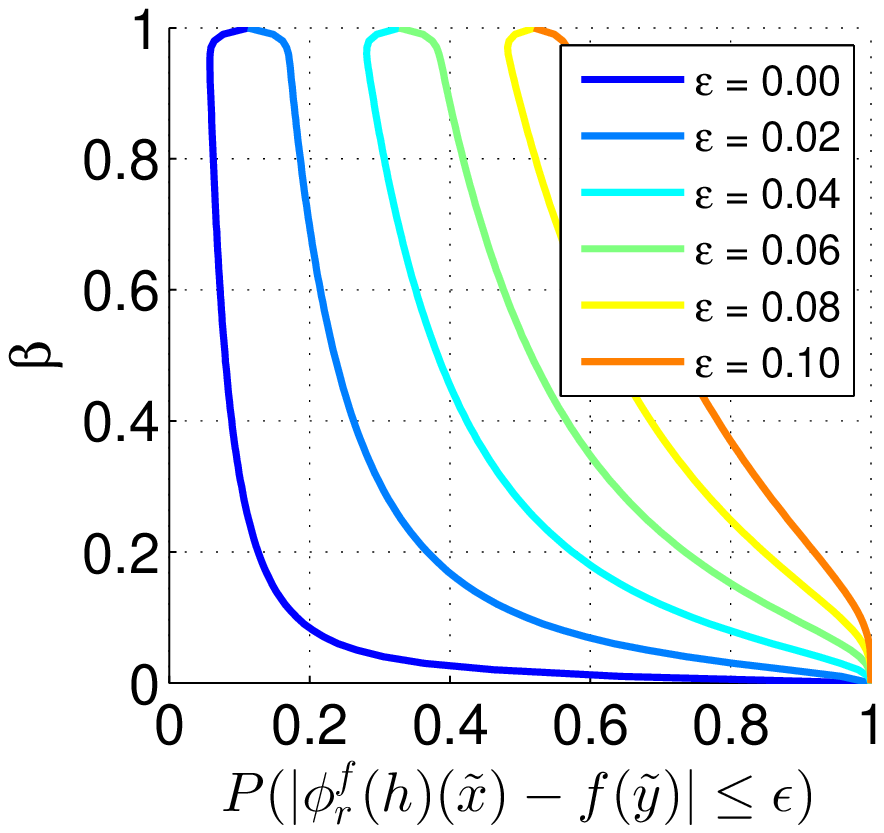}}
\subfigure[$u(r,\epsilon)$]
{\includegraphics[width = 3.4cm]{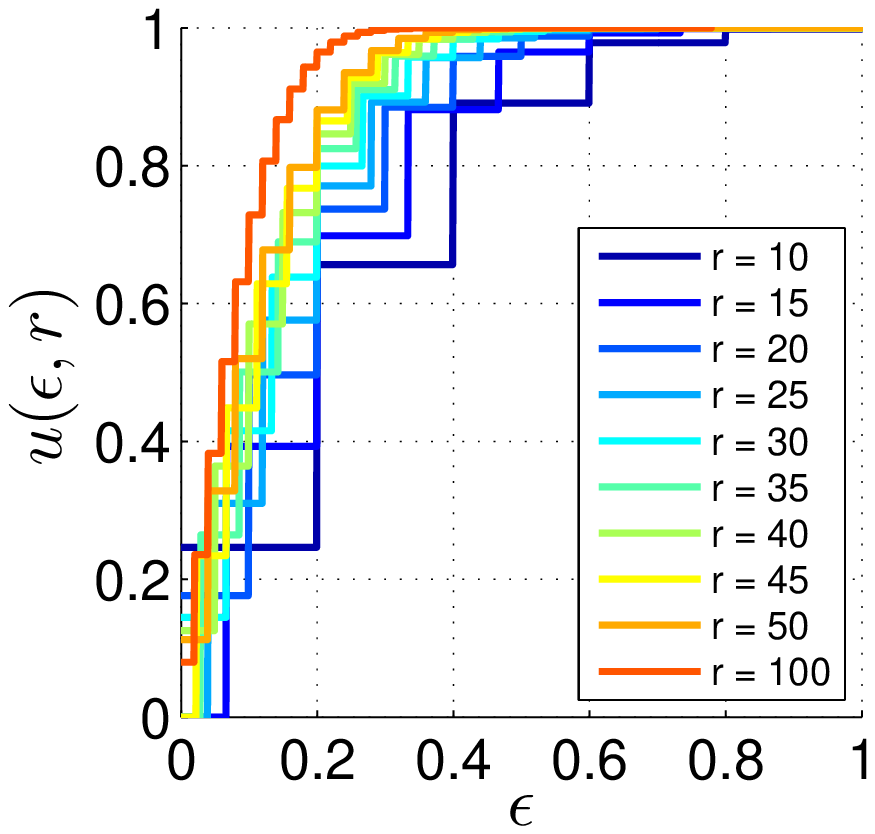}}
\vspace{-0.35cm}
\caption{(a)-(c): Relationship of the probability of making wrong label prediction, $\beta : = \mathbb{P}(h(x) \neq y)$, and the probability of bag proportion prediction error is small than $\epsilon$, $\mathbb{P} \left( |\phi_r^f(h)(\tilde{x}) -  f(\tilde{y})| \leq \epsilon \right)$, under the assumption that the instances are drawn \emph{iid}, and the prior can be matched by the hypothesis, \emph{i.e.}, $\mathbb{P} (h(x) = 1) = \mathbb{P}( y=1 )$.
$\beta$ is a monotonically decreasing function of  $\mathbb{P} \left( |\phi_r^f(h)(\tilde{x}) -  f(\tilde{y})| \leq \epsilon \right)$, if $\mathbb{P} \left( |\phi_r^f(h)(\tilde{x}) -  f(\tilde{y})| \leq \epsilon \right) \in (u(r,\epsilon), 1]$. $u(r,\epsilon)$ is shown in (d). Larger $r$ and larger $\epsilon$ will result in larger $u(r,\epsilon)$.} 
\label{fig:iid}
\vspace{-0.3cm}
\end{figure*}

We can instead use the generative assumption to provide stronger results: under some mild assumptions, the probability of making wrong label prediction $\beta := \mathbb{P} (h(x) \neq y)$ can be expressed as a function of 
$\mathbb{P} \left( \left| \phi_r^f(h)(\tilde{x}) -  f(\tilde{y}) \right| \leq \epsilon \right)$. 
Let's first consider a special case, when \emph{all} the instances are drawn \emph{iid} over a distribution of the instances. This is a special case of the generative model mentioned above with only one component in $\mathcal{D}$. In addition, we assume that the prior of the instances can be matched by the hypothesis, 
\emph{i.e.}, $\mathbb{P} (h(x) = 1) = \mathbb{P}( y=1 )$. This assumption is not too restrictive, 
as the small empirical bag proportion error already implies that the priors are approximately matched, and for most learning models, we can adjust the hypothesis by a bias term to match the empirical prior estimated on the training data.
In such case, we can express  $\mathbb{P} \left( \left| \phi_r^f(h)(\tilde{x}) -  f(\tilde{y}) \right| \leq \epsilon \right)$ analytically as a function of $\beta$: 
\begin{small}
\begin{align}
\mathbb{P} \left( \left| \phi_r^f(h)(\tilde{x}) -  f(\tilde{y}) \right| \leq \epsilon \right)  =  \theta_1^r \sum_{i = 0}^r {{r} \choose {i}} 
\theta_2 ^{i}
\Big( \mathcal{F}(i + \lfloor \epsilon r \rfloor; r-i, \theta_2) - \mathcal{F}(i - \lfloor \epsilon r \rfloor -1; r-i, \theta_2)\Big), \nonumber
\end{align}
\end{small}
where $\lfloor \cdot \rfloor$ is the floor operator,  $\theta_1 = (2-\beta)/2$, $\theta_2 = \beta/(2-\beta)$, $0 < \beta < 1$, $0 < \epsilon < 1$ and $\mathcal{F}$ is the CDF of binomial distribution.

As what we actually want is to bound $\beta$ based on $\mathbb{P} \left( \left| \phi_r^f(h)(\tilde{x}) -  f(\tilde{y}) \right| \leq \epsilon \right)$, we show the ``inverse'' function in
Figure \ref{fig:iid}. 
From the curves, we see that $\beta$ is a monotonically decreasing function of $\mathbb{P} \left( \left| \phi_r^f(h)(\tilde{x}) -  f(\tilde{y}) \right| \leq \epsilon \right)$, when 
$\mathbb{P} \left( |\phi_r^f(h)(\tilde{x}) -  f(\tilde{y})| \leq \epsilon \right) \in (u(r,\epsilon), 1]$. $u(r,\epsilon)$ is shown in Figure \ref{fig:iid} (d). In other words, the instance label error can be controlled by the bag proportion error, when the later is small. The results are of course the ``tightest'' under the above assumptions.

One important observation is that the curves are independent of $\mathbb{P}( y=1 )$. Therefore, the analytical results can be directly applied to the case when the instances are conditionally independent given the bag, under the assumption that $\mathbb{P}_{(x,y) \sim D'_i} (h(x) = 1) = \mathbb{P}_{(x,y) \sim D'_i}(y=1), \forall i$. 

Our additional observations of Figure \ref{fig:iid} are summarized below. From (b), the growth of $\beta := \mathbb{P}(h(x) \neq y)$ is approximately linear to the bag size $r$, when $r$ is sufficiently large, and $\mathbb{P} \left( |\phi_r^f(h)(\tilde{x}) -  f(\tilde{y})| \leq \epsilon \right)$ is close to 1. From Figure \ref{fig:iid} (c), with large bag size ($r=50$), the growth of $\beta$ is approximately quadratic to the proportion prediction error $\epsilon$, when $\mathbb{P} \left( |\phi_r^f(h)(\tilde{x}) -  f(\tilde{y})| \leq \epsilon \right)$ is close to 1.

\subsection{Instances Are Drawn by Independent Bernoulli Variables}
\label{subsec:kappa}
In this section, we focus on an alternative model, which essentially picks instances independently from a finite set of instances to build the bags. This setting is relevant to most real-world applications. For example, there are always a finite set of persons in a political survey. 
Our intuition is that if the bags ``cover'' the space of the instances well, the instance labels (of all possible instances) are relatively easy to be recovered.

Formally, we consider the following model for bag generation. 
Let $X = \{x_{1},...,x_{n}\}$ be the set of \emph{all} instances. For each instance $x_{i}$, define an Bernoulli variable $\kappa_{i}$ that determines whether the instance  will be picked, \emph{i.e.}:
\[ \kappa_{i} = \left\{ 
  \begin{array}{l l}
    1 & \quad \text{if $x_{i}$ was selected to the bag,}\\
    0 & \quad \text{otherwise.}
  \end{array} \right.\]
Let $p_{i} = \mathbb{P} (\kappa_{i}=1)$. Note that $\kappa_{i}$ are \emph{independent but not necessarily identically distributed}. Denote $\kappa=(\kappa_{1},..,\kappa_{n})$. We call the above model the \textit{$\kappa$-model}. Note that in the $\kappa$-model, the bag size is variable instead of being fixed. The proof of the results is shown in the appendix.%
\begin{theorem}
\label{thm:kappa}
Let $X = \{x_{1},\cdots,x_{n}\}$ be the set all the instances. 
Let $\max_{i} p_{i} = \mathcal{O}(\frac{1}{n})$. 
Assume that $\min_{i} p_{i} \geq R/n$ for some $R>0$.
Let $D$ be a distribution over bags such that drawing a bag from $D$ is by drawing a bag from the $\kappa$-model. 
Let $h$ be a hypothesis such that 
$\mathbb{P}_{(\tilde{x}, \tilde{y}) \sim D} (|\phi_{r(\tilde{x})}^f(h)(\tilde{x})-f(\tilde{y})| \leq \epsilon ) \geq 1 - \delta$, where $r(\tilde{x})$ is the size of $\tilde{x}$. Let $q = \frac{\epsilon^{2}(\sum_{i=1}^{n} p_i)^{2}}{n \min_{i} p_i(1-p_i)}$.
Then, for small enough $\delta>0$ and large enough $R$, $h$ misclassifies at most
$\mathcal{O}(qn)$ instances of $X$.
\end{theorem}
The following corollary shows the result when the probabilities of the instances being drawn are uniform, and equal to $\mathcal{O}(\frac{1}{n})$. In such case, the instances are \emph{iid}, and the bags very well ``cover'' $X$. This is similar to one case studied in Section \ref{subsec:conditionaliid} when all instances are drawn \emph{iid}.
\begin{corollary}
\label{coro:kappa}
Let $X = \{x_{1},...,x_{n}\}$ be the set all the instances. 
Let $p_{1} = \cdots = p_{n} = p =  \mathcal{O}(\frac{1}{n})$.
Denote by $\hat{r}$ the expected size of the bag, $\hat{r}=np$. 
Let $D$ be a distribution over bags such that drawing a bag from $D$ is by drawing a bag from the above $\kappa$-model. 
Let $h$ be a hypothesis such that 
$\mathbb{P}_{(\tilde{x}, \tilde{y}) \sim D} (|\phi_{r(\tilde{x})}^f(h)(\tilde{x})-f(\tilde{y})| \leq \epsilon ) \geq 1 - \delta$.
Then for small enough $\delta$ and large enough $\hat{r}$, 
$h$ misclassifies at most 
$\mathcal{O}(\epsilon^{2}\hat{r}n)$ instances of $X$.
\end{corollary}

Similar to Section \ref{subsec:conditionaliid}, the model has an analogous ``conditionally-independent'' version, where we have a probability distribution on the set of sequences $\kappa$. In this setting we first choose a sequence $\kappa$ and then use it to generate a particular bag. All the results we give above easily translate to this more general setting. In addition, based on our proof in the appendix, it is also possible to generalize the result to cases where dependencies between small subsets of instances exist.

\section{Extending the Results to Variable Bag Size}
\label{sec:bag_size}
For simplicity in our analysis, except Section \ref{subsec:kappa}, we assumed that all the bags were of size $r$. In fact, all our results can be easily generalized to variable bag size. 
For the result on the bag proportion error, Theorem \ref{thm:bag} holds immediately by replacing $r$ with the \emph{average bag size in training} $\bar{r}$. This is based on the fact that the covering number bound holds (shown in the proof of Theorem \ref{thm:bag} in the appendix) with average bag size \cite{sabato2012partial}. 
For results on the instance label error, we can assume that the bag size $r$ is generated from some fixed probability distribution independent of the instances. 
Lemma \ref{lemma:instance} holds for any bag size. Theorem \ref{thm:instance} holds immediately by replacing $r$ with the \emph{expected bag size} $\hat{r}$, following the proof described in the appendixappendix. We can further bound the \emph{expected bag size} $\hat{r}$ by the \emph{average bag size in training} $\bar{r}$ based on Chernoff bound: for any $t > 0$, with probability at least $1 - e^{-2mt^2}$, $\hat{r} > \bar{r} - t$, where $m$ is the number of training bags. In addition, our analysis with the generative assumptions (Section \ref{subsec:conditionaliid}) holds for any bag size.

\section{Applications}
\label{sec:applications}

\textbf{LLP for privacy preserving learning algorithms.}
Data used by machine learning algorithms is often very sensitive and therefore
versions of the machine learning algorithms that achieve some level of privacy are of much interest \cite{choro2, dwork2006differential}.
LLP can be very useful to provide data privacy. In a nutshell, the LLP learner, only having access to the label proportions, suffices to
construct good-quality machine learning structures. We provide preliminary discussions in the appendix.

\textbf{Justifying LLP algorithms.}
Among the existing algorithms, our analysis is well aligned with algorithms that implicitly utilize the EPRM framework. 
%
One algorithm in this category is $\propto$SVM \cite{Yu13_3}, which has been shown to give good performance.
The objective of $\propto$SVM is to find a large-margin classifier consistent with the label proportions, with the help of latent labels. The analysis of Section \ref{sec:iidbag} can be generalized to justify $\propto$SVM using margin-based analysis \cite{zhang2002covering}.

\textbf{Learning with population proportions.}
Consider the scenario of modeling voting behavior based on government released statistics: the government releases the population proportion (\emph{e.g.} 62.6\% in New York voted for Obama in 2012 election) of each location, and we only have a subset of randomly sampled instances for each location. Can LLP be applied to correctly predict labels of the individuals?
In such a scenario, EPRM can only minimize the proportion error in terms of the population proportions, because the actual proportions of the sampled subsets are not available. 
We can assume that a bag is formed by randomly sampling $r$ instances $\{(x_i, y_i)\}_{i=1}^r$ from a location with a true (population) proportion $p^*$, which is released. 
The Chernoff bound ensures that the sampled proportion is concentrated to the released population proportion: \emph{when $r \geq \ln(2/\delta)/(2\epsilon^2)$, with probability at least $1-\delta$,
$|f(\tilde{y})-p^*| \leq \epsilon$.} Therefore, with enough training bags, and enough samples per bag, the generalization error can be bounded.

\section{A Case Study: Predicting Income based on Census Data}
\label{sec:exp}
\vspace{-0.1cm}

\begin{figure*}[t]
\centering
\subfigure[IID]
{\includegraphics[width = 3.4cm]{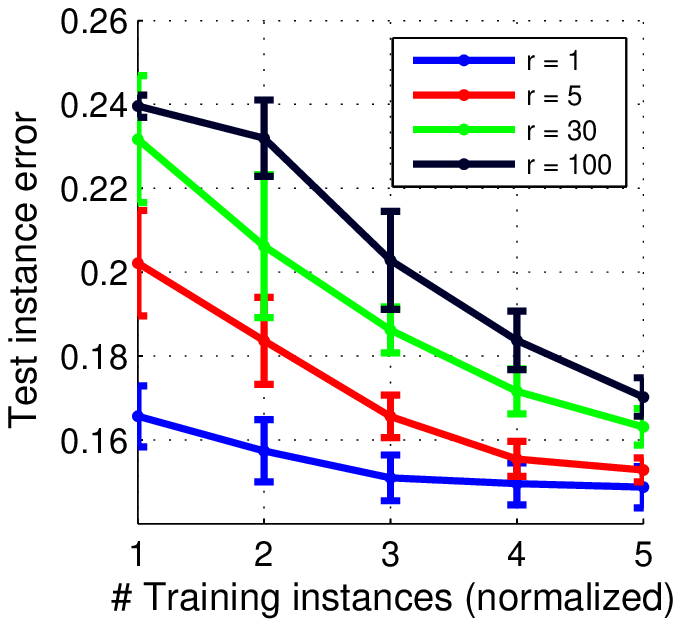}}
\hspace{-0.2cm}
\subfigure[Occupation]
{\includegraphics[width = 3.4cm]{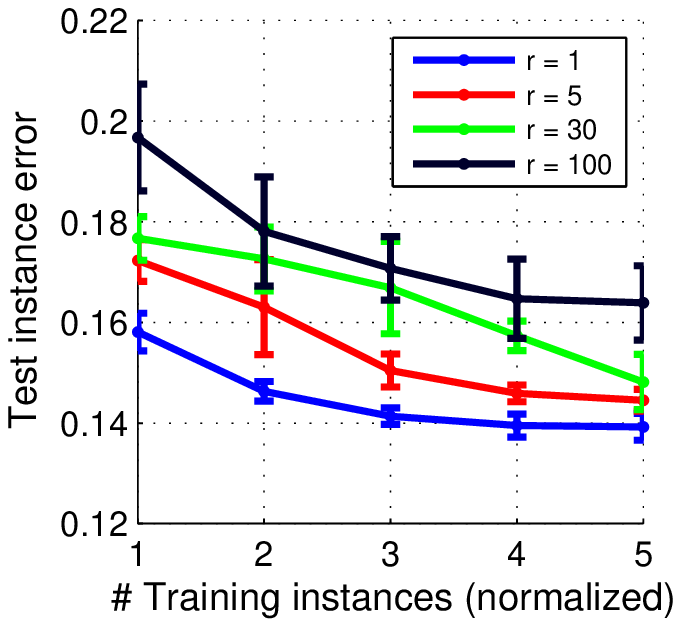}}
\subfigure[Education]
{\includegraphics[width = 3.4cm]{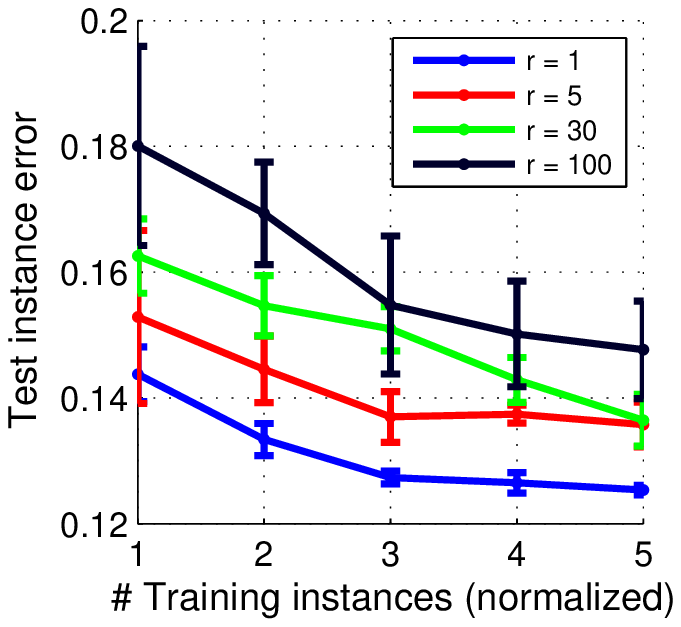}}
\subfigure[Race]
{\includegraphics[width = 3.4cm]{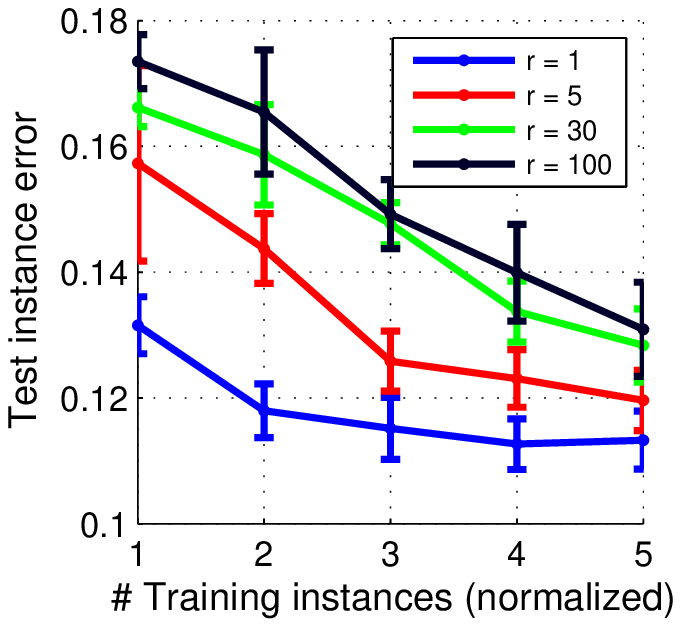}}
\vspace{-0.35cm}
\caption{Predicting income based on census data. 
(a) All the instances are \emph{iid}. (b)-(d) The instances are conditionally independent given the bag, with title of each figure as its grouping attributes. The number of training instances is equal-spaced in log-scale, with the smallest number 500, and the largest number 50,000.}
\label{fig:exp_census}
\vspace{-0.4cm}
\end{figure*}

So far we have provided formal guarantees that under some conditions, 
labels of individual instances can be learned or recovered.
In this section, we conduct a case study to demonstrate the feasibility of LLP on real-world data. The task is to predict individual income based on census data, and label proportions. 
We use a dataset which covers a subset of the 1994 census data\footnote{\url{http://archive.ics.uci.edu/ml/datasets/Adult}}. It contains 32,561 instances (individual persons), each with 123 binary attributes about education, marital status, sex, occupation, working hours per week \emph{etc}. Each individual has a binary label indicating whether his/her \emph{income $>$ $50k$}.
We consider this label as sensitive information, for which we only know its proportions on some bags of people.
Next we will show the feasibility of LLP based on different ways of forming the bags. 
We first divide the dataset to use 80\% of the instances for training, and 20\% for testing. 
For all the experiments below, the bags are formed on the 80\% training data. The proportion of each training bag is computed based on the ground-truth labels. 
The experiments are based on $\propto$SVM \cite{Yu13_3} with linear kernel, and $\ell_1$ loss. We use the algorithms proposed in \cite{rueping2010svm} and \cite{quadrianto2009estimating} as the initialization methods for the alter-$\propto$SVM solver. 
The parameters, $C \in [0.1, 1, 10]$, $C_p \in [0.01, 0.1, 1]$, are tuned on the bag proportion prediction error by performing cross validation. 
We report mean and standard deviation of error on the test data based on 5 runs.

\textbf{All instances are drawn \emph{iid}.}
We first simulate a simple case where all the instances are drawn \emph{iid} from a distribution of instances. 
We achieve this by assuming that all the instances of the dataset form a ``distribution'' of individuals. For all the bags (with $r$ instances), each instance is drawn by randomly sampling the whole training set with replacement. Figure \ref{fig:exp_census} (a) shows the experimental results. 
Based on the results, more instances (bags) lead to lower test error. The relatively low test error demonstrates the feasibility of LLP under such settings. Fewer training bags (and larger training bag size) generally result in larger test error variance, as the algorithm is more likely to converge to worse local solutions.

\textbf{Instances are conditionally independent given bag.}
We simulate the case by a ``hierarchical bag generation'' process. The individuals are first grouped by a \emph{grouping attribute}.
For example, if the grouping attribute is ``occupation'', 
the individuals are groups into 15 groups, corresponding to the 15 occupations. 
Each group is assigned a prior, which is simply uniform in this experiment. 
To generate a bag of $r$ instances, we first pick a group based on the priors, and then perform random sampling with replacement $r$ times in the selected group.
Figure \ref{fig:exp_census} (b)-(d) show the experiment results. The trend of the learning curves is the same as in Figure \ref{fig:exp_census} (a), showing that LLP is indeed possible in such a setting. 

\begin{table*}[t]
\begin{center}
\begin{small}
\begin{tabular}{l|l|l|l|l|l}
\hline
Grouping Attribute        &      Native Country&   Education      &  Occupation       &         Relationship         & Race\\
\hline
Number of Bags            &       41            &       16          &    15            &             5         & 5     \\
\hline   
$\propto$SVM Error \%    & 18.75 $\pm$ 0.25   &  19.61 $\pm$ 0.10  & 18.19 $\pm$ 0.16  &    18.59 $\pm$ 0.82   &   24.02 $\pm$ 0.15  \\
\hline
Baseline Error \%  & 24.02 $\pm$ 0.56   &  22.29 $\pm$ 0.55  &  24.28 $\pm$ 0.28 &   24.19 $\pm$ 0.72   & 24.28 $\pm$ 0.40     \\
\hline
\end{tabular}
\end{small}
\end{center}
\vspace{-0.35cm}
\caption{Error on predicted income based on census in a real-world setting. }
\label{table:income}
\vspace{-0.5cm}
\end{table*}

\textbf{A challenging real-world scenario.}
For real-world applications, the bags are predefined rather than ``randomly'' generated.
In this experiment, 
we simply group the individuals defined by different grouping attributes. Different from the setting in the former section, the whole group is treated as a single bag, without further sampling process.
Table \ref{table:income} shows the performance of LLP. 
The ``Baseline'' result is formed by the following: a new instance is predicted positive if the training bag with the same grouping attribute has proportion larger than 50\%; otherwise it is predicted as negative. For example, it predicts a person with elementary education as negative, as the label proportion of elementary education group is less than 50\%. 
For most experiments, this result is similar to assigning +1 to all test instances, because most training bags are with proportion larger than 50\%. 
This scheme provides performance gain for ``education'', as the label proportion for individuals with very low education is also very low. 
We find that for most grouping attributes, the performance of LLP significantly improves over the baseline. This is also true for some cases where the number of bags is very small. 
We do observe that for certain way of forming the bags, \emph{e.g.}, by Race, the improvement is not that big. This is due to the fact that  the instance distributions of the bags are very similar, and therefore most of the bags are redundant for the task. 

\section{Conclusion}
This paper proposed a novel two-step analysis to answer the question whether the individual labels can be learned when only the label proportions are observed. We showed how parameters such as bag size and bag proportion affect the bag proportion error and instance label error.
Our first result shows that the generalization error of bag proportions is only mildly sensitive to the size of the bags. Our second result shows that under different mild assumptions, a good bag proportion predictor guarantees a good instance label predictor. 
We have also demonstrated the feasibility of LLP based on a case study. 
As the future work, data dependent measure, \emph{e.g.}, Rademacher complexity \cite{bartlett2003rademacher}, may lead to tighter bound for practical use. Some alternative tools, \emph{e.g.}, sample complexity results of learning $\{0,..., n\}$-valued functions \cite{haussler1995generalization}, can be used for analyzing the generalization error of bag proportions. The failure cases of LLP worth further study as they can be utilized to protect sensitive personal information when releasing label proportions.

\section{Appendix}
\subsection{Proof Sketch of Theorem 1}

One important tool used in the proof is the lemma below bounding the covering number of bag proportion hypothesis class $\phi_r^f(\mathcal{H})$ by the covering number of the instance hypothesis class $\mathcal{H}$.

\begin{Lemma} \cite{sabato2012multi, sabato2012partial}
Let $r \in \mathbb{N}$ and suppose $f: \mathbb{R}^r \rightarrow \mathbb{R}$ is $\alpha$-Lipschitz w.r.t. 
the infinity norm, for some $\alpha_f >0$. Let  $\gamma > 0$, $p \in [1, \infty]$, and $\mathcal{H} \in \mathbb{R}^\mathcal{X}$. For 
any $m \geq 0$,
\begin{equation*}
\mathcal{N}_p(\gamma, \phi_r^f(\mathcal{H}), m) \leq \mathcal{N}_{p}(\frac{\gamma}{\alpha_f r^{1/p}}, \mathcal{H}, rm).
\end{equation*}
\vspace{-0.1cm}
\label{lemma:sabato}
\end{Lemma}
Covering number \cite{anthony2009neural} can be seen as a complexity measure on real-valued hypothesis class. The larger the covering number, the larger the complexity. Another lemma we use is the uniform convergence for real function class.
\begin{Lemma}\cite{anthony2009neural}.
Let $\hat{\mathcal{Y}},\mathcal{Y} \subseteq \mathbb{R}$, $\mathcal{G} \subseteq \hat{\mathcal{Y}}^\mathcal{X}$, and $L: \hat{\mathcal{Y}} \times \mathcal{Y} \rightarrow [0, 1]$, such that $L$ is Lipschitz in its first argument with Lipschitz constant $\alpha_L > 0$. Let $D$ be any distribution on $\mathcal{X} \times \mathcal{Y}$. Then for any $0< \epsilon < 1$ and $g \in \mathcal{G}$:
\begin{align*}
& \mathbb{P}_{S \sim D^m} \left( \sup_{g \in \mathcal{G}} \left| \text{er}_D^L(g) - \text{er}_S^L(g) \right| \geq \epsilon   \right) \leq 
4 \mathcal{N}_1 (\epsilon/(8\alpha_L), \mathcal{G}, 2m) e^{-m\epsilon^2 / 32},
\end{align*}
in which $er_S^L(g) = \frac{1}{|S|}\sum_{x\in S} L(g(x), y)$, $er_D^L(g) = \mathbb{E}_{x \sim D} L(g(x), y)$.
\label{lemma:uniform}
\end{Lemma}
Based on the definition of covering number: $\mathcal{N}_1(\epsilon, W, m) \leq \mathcal{N}_\infty(\epsilon, W, m)$. 
Applying Lemma \ref{lemma:sabato}:
\begin{align}
 4 \mathcal{N}_1 (\epsilon/(8\alpha_L), {\mathcal{H}}, 2m) e^{-m\epsilon^2 /32} \nonumber 
\leq & 4 \mathcal{N}_\infty (\epsilon/(8\alpha_L), {\mathcal{H}}, 2m) e^{-m\epsilon^2 /32} \nonumber \\
\leq & 4 \mathcal{N}_\infty (\epsilon/(8\alpha_L\alpha_f), \mathcal{H}, 2rm) e^{-m\epsilon^2 /32}.
\label{eq:cover}
\end{align}
Also, the proportion generation function $f$ in our case is 1-Lipschitz with the infinity norm. And the loss functions $L$ we are considering is $1$-Lipschitz.

Based on the definition of covering number, for $\mathcal{H} \subseteq \{-1, 1\}^{\mathcal{X}}$, for any $\epsilon < 2$, $\mathcal{N}(\epsilon, \mathcal{H}_{| x_1^m}, d_\infty) = |\mathcal{H}_{| x_1^m}|$. Thus,
\begin{equation*}
\mathcal{N}_\infty(\epsilon, \mathcal{H}, m) = \Pi_\mathcal{H}(m).
\end{equation*}

Refer to \cite{anthony2009neural} for the definition of \emph{restriction} $\mathcal{H}_{| x_1^m}$ and \emph{growth function} $\Pi_\mathcal{H}(m)$. In addition, we have the following lemma to 
to bound the growth function by VC dimension of the hypothesis class.
\begin{Lemma} \cite{sauer1972density}
Let $\mathcal{G} \subseteq \{-1,1\}^\mathcal{X}$ with $\text{VC}(\mathcal{G}) = d \leq \infty$. For all $m \geq d$, 
$\Pi_{\mathcal{G}}(m) \leq \sum_{i=0}^d {m\choose i} \leq \left( \frac{em}{d} \right)^d$.
\end{Lemma}
Let $d = VC(\mathcal{H})$. By combining the above facts, and $0 < \epsilon < 1$, (\ref{eq:cover}) leads to
\begin{align*}
4\Pi_\mathcal{H}(2rm) e^{-m\epsilon^2 /32} \leq 4\left( \frac{2erm}{d} \right)^{d} e^{-m\epsilon^2 /32}.
\end{align*}
Therefore, with probability at least $1-\delta$,
\begin{align*}
& \text{er}_D^L(f) \leq \text{er}_S^L(f) + 
 \left( \frac{32}{m}(
d\ln(2emr/d) + \ln(4/\delta))
\right)^{1/2} \nonumber \\
& \Leftarrow m \leq \frac{32}{\epsilon^2} (d\ln m + d \ln(2er/d) + \ln (4/\delta)).
\end{align*}
Since $\ln x \leq a x - \ln a -1$ for all $a,x > 0$, we have
\begin{align*}
& \Leftarrow \frac{32d}{\epsilon^2} \ln m \leq  \frac{32d}{\epsilon^2} \left(\frac{\epsilon^2}{64d}m + \ln\left( \frac{64d}{\epsilon^2}\right) \right) 
\leq  \frac{m}{2} + \frac{32d}{\epsilon^2}\ln\left( \frac{64d}{e\epsilon^2} \right). \\
& \Leftarrow  m \geq \frac{m}{2} + \frac{32}{\epsilon^2}(d\ln(128r/\epsilon^2) + \ln(4/\delta)). \\
& \Leftarrow  m \geq \frac{64}{\epsilon^2}(2d\ln(12r/\epsilon) + \ln(4/\delta)). 
\end{align*}

\subsection{Appendix of Section 6.1}

\textbf{Proof of Lemma 1}

 Let $\tilde{x}=(x_{1},...,x_{r})$ be a bag. Assume that $| \phi_r^f(h)(\tilde{x}) - f(\tilde{y}) | \leq  \epsilon$.
Denote $\mathcal{A}_{1}=\{i \in \{1,...,r\}:h(x_{i})=+1 \wedge y_{i} = -1\}$, $\mathcal{A}_{2}=\{i \in \{1,...,r\}:h(x_{i})=-1 \wedge y_{i} = +1\}$, $\mathcal{A}_{3}=\{i \in \{1,...,r\}:h(x_{i})=+1 \wedge y_{i} = +1\}$,
$\mathcal{A}_{4}=\{i \in \{1,...,r\}:h(x_{i})= -1 \wedge y_{i} = -1\}$.
We have: $\phi_r^f(h)(\tilde{x})=(|\mathcal{A}_{1}|+|\mathcal{A}_{3}|)/r$,
$ f(\tilde{y})  = (|\mathcal{A}_{2}|+|\mathcal{A}_{3}|)/r$.

Thus, from the assumption: $| \phi_r^f(h)(\tilde{x}) - f(\tilde{y}) | \leq  \epsilon$, we have: 
$||\mathcal{A}_{1}|-|\mathcal{A}_{2}|| \leq \epsilon r$.
Assume that $\tilde{x}$ is \emph{$(1-\eta)$-pure}. Without loss of generality we can assume that 
$|\mathcal{A}_{1}|  + |\mathcal{A}_{4}| \leq \eta r$. This implies  that $|\mathcal{A}_{1}| \leq \eta r$. Thus we also have that $|\mathcal{A}_{2}| \leq (\eta + \epsilon)r$.
Hypothesis $h$ correctly classifies $|\mathcal{A}_{3}| + |\mathcal{A}_{4}|$ instances of the bag $\tilde{x}$.
From what we have just derived, we conclude that $h$ correctly classifies at least $(1-2\eta-\epsilon)r$ instances of the
bag $\tilde{x}$.
So far in the analysis above we assumed that: $| \phi_r^f(h)(\tilde{x}) - f(\tilde{y}) | \leq  \epsilon$ and
 $\tilde{x}$ is \emph{$(1-\eta)$-pure}. From the statement of the theorem we know that the former happens with
probability at least $1-\delta$ and the latter with probability at least $1-\rho$. 
Thus, by the union bound, we know that with probability at least $1-\delta-\rho$, $h$ classifies
correctly at least $(1-2\eta-\epsilon)$ instances of the bag drawn from the distribution $\mathcal{D}$.

\textbf{Proof of Theorem 2}

Denote $\theta = 2\eta + \epsilon$, $p=\delta + \rho$.
Let $Z$ be a random variable defined as follows:
\[ Z = \left\{ 
  \begin{array}{l l}
    (1-\theta)r & \quad \text{with probability $1-p$,}\\
    0 & \quad \text{with probability $p$}
  \end{array} \right.\]
Denote by $Z_{i}$ for $i=1,2,...,m$ the $i^{th}$ independent copy of $Z$.
Lets $Z_{sum}=Z_{1}+...+Z_{n}$.
Note that, according to Lemma 1, $Z$ is the lower bound on the number of instances from $m$ bags that are correctly classified
We have $\mu = \mathbb{E} Z_{sum}=nr(1-p)(1-\theta)$. From Chernoff's inequality we immediately get: for $0 < \tau < 1$,
$\mathbb{P}(Z_{sum} < (1-\tau)\mu) \leq e^{-\frac{\tau^{2}}{2}\mu}$.
Thus we conclude that with probability at least $1 - e^{-\frac{\tau^{2}}{2}nr(1-p)(1-\theta)}$ the hypothesis $h$
classifies correctly at least a fraction $(1-\tau)(1-p)(1-\theta)$ of all $nr$ instances coming in $n$ bags.

\textbf{Proof of Lemma 2}

Assume that the bags are selected in such a way that first $\eta r$ instances have label $+1$,
next $\eta r$ have label $-1$ and last $(1-2\eta) r$ instances have label $+1$.
Assume furthermore that the hypothesis $h$ satisfies: $h(x_{i}) = - y_{i}$ for every instance $x_{i}$ from the list
of first $2\eta r$ instances of the bag and satifies: $h(x_{i}) = y_{i}$ for last $(1-2\eta)r$ instances from the list.
Then of course the classifier always misslassifies first $2\eta$ instances of each bag but predicts bag proportions
with $0$ error.

\subsection{Appendix of Section 6.3}

\textbf{Proof of Theorem 3}

The theorem follows from the following more general technical theorem that we will prove first. Note that we treat the labels as 0/1 for simplicity in this section.

\begin{theorem}
\label{main_technical_theorem}
Let $(y_{1},\cdots,y_{n})$ and $(h_{1},\cdots,h_{n})$ be two sequences of $0$s and $1$s.
Denote $\eta = \frac{\sum_{i=1}^n|y_{i}-h_{i}|}{n}$. Let $0 < \epsilon < 1$.
Let $p=\max_{i} p_{i}$. 
Let $v= \min_{i=1,...,n} Var(\kappa_{i})$.
Let $\rho(x) = \sum_{t=\sqrt{x \eta n v}}^{n} 
\frac{e^{t(\log(np)+1)}}{e^{(t+\frac{1}{2})t}}$.
Assume that: $x \rho(x) \rightarrow 0$ as $x \rightarrow \infty$ and $\int_{x}^{\infty} \rho(x) \rightarrow 0$ as $x \rightarrow \infty$.
Assume furthermore that $\eta \geq \frac{\Delta \epsilon^{2}(\sum_{i=1}^{n}E\kappa_{i})^{2}}{n v}$
for some constant $\Delta>0$.
Then the following is true for $\Delta$, $n$ big enough, and $\sum_{i=1}^n Var \big( \kappa_i (y_i - h_i) \big)$ big enough:
$$\mathbb{P}_{\kappa=(\kappa_{1},...,\kappa_{n})}\left( \left| \sum_{i=1}^{n} \kappa_{i}(y_{i}-h_{i}) \right|  > \sum_{i=1}^{n}\kappa_{i} \epsilon \right) > \xi$$
for some constant $\xi(\Delta)$ that does not depend on $\eta$ and $\epsilon$.
\end{theorem}

\textbf{Proof}

Denote $X_{i} = \kappa_{i}(y_{i}-h_{i})$, $X = \sum_{i=1}^{n} X_{i}$, $Z_{i}=X_{i}-EX_{i}$, $Z = \sum_{i=1}^{n} Z_{i}$
and $Y=(X-EX)^{2}$. Notice that we have: $EY = Var(X) = \sum_{i=1}^{n} Var(X_{i})$.
First we want to find a lower bound on the probability $p_{1}=\mathbb{P}_{\kappa=(\kappa_{1},...,\kappa_{n})}[|\sum_{i=1}^{n} \kappa_{i}(y_{i}-h_{i})| > C \sqrt{Var(Z)}]$ for some constant $C>0$ (we will show later how to choose $C$).

Let us consider two cases. In the first case we have: $|EX| \leq C\sqrt{Var(Z)}$.
So it only suffices to find a lower bound on the probability $p_{2}=\mathbb{P}[|X-EX| > 2C\sqrt{Var(Z)}]$, because trivially:
$p_{1} \geq p_{2}$.
Let $\alpha, \beta > 0$ ($\alpha < \beta$) be two positive constants (we will see later how to choose the values of $\alpha$ and $\beta$).
To find a lower bound on $p_{2}$ we partition the probability space $\Omega$ into three subspaces: 
$\Omega_{1}, \Omega_{2}, \Omega_{3}$. 
Let us define: $\Omega_{1} = \{Y < \alpha Var(Z)\}$, $\Omega_{2} = \{\alpha Var(Z) \leq Y \leq \beta Var(Z)\}$,
$\Omega_{3} = \{Y > \beta Var(Z)\}$.
We trivially have: $EY \leq \mathbb{P}[A]E(Y|A)+\mathbb{P}[B] \beta Var(Z) + \mathbb{P}[C]E(Y|C)$.
Let us first consider the expression: $\mathbb{P}[C]E(Y|C)$ and lets fix some $\alpha>0$.
We want to show that for $\beta$ large enough we have: $\mathbb{P}[C]E(Y|C) < \alpha Var(Z)$.
So we want to prove that for $\beta$ large enough the following is true:
$E[YI\{Y>\beta Var(Z)\}] < \alpha Var(Z)$. The latter inequality is equivalent to the following one:
$E[\frac{Y}{Var(Z)} I\{\frac{Y}{Var(Z)} > \beta\}] < \alpha$.
Let us denote $D_{n}=\frac{Z}{\sqrt{Var(Z)}}$. We want to show that $E[D_{n}^{2}I\{D_{n}^{2} > \beta\}] < \alpha$.
Let us first bound the expression $\mathbb{P}[|D_{n}| > x]$ for an arbitrary $x>0$.
Notice that $\mathbb{P}[D_{n} > x] = \mathbb{P}[\sum_{i=1}^{n}(X_{i}-EX_{i}) \geq x\sqrt{\eta n v}] $.
Now notice that each $X_{i}$ is either $0$, $1$ or $-1$ and that different $X_{i}$ are independent.
Thus we have: $\mathbb{P}[D_{n} > x] \leq {n \choose t} p^{t}$, where $t = x\sqrt{\eta n v}$.
Thus we get that: 
\begin{equation*}
\mathbb{P}[D_{n} > x] \leq \frac{(np)^{t}}{t!} = O(\frac{e^{t\log(np)+t}}{e^{(t+\frac{1}{2})\log(t)}})
\end{equation*}
where the last relation follows immediately from Stirling's formula.
Thus we have: 
\begin{equation*}
E[\frac{Y}{Var(Z)} I\{\frac{Y}{Var(Z)} > \beta\}]  = \beta \mathbb{P}[D_{n}^{2} > \beta] + 
\int_{\beta}^{\infty} \mathbb{P}[D_{n}^{2}>x] \mathrm{d}x = o(1)
\end{equation*}
 (\emph{i.e.} the expression goes to $0$ as $x \rightarrow \infty$), 
where the last relation follows from our assumptions on 
$\rho(x)$ and the upper bound on the value of $\mathbb{P}[D_{n} > x]$ that we have just derived.
We can conclude that for $\beta$ large enough we have: $\mathbb{P}[C]E(Y|C) < \alpha Var(Z)$.
Now, if we go back to the formula: $EY \leq \mathbb{P}[A]E(Y|A)+\mathbb{P}[B] \beta Var(Z) + \mathbb{P}[C]E(Y|C)$
and notice that $EY = Var(Z)$, then from the fact that $\mathbb{P}[C]E(Y|C) < \alpha Var(Z)$, we obtain:
$\mathbb{P}[B] \geq \frac{1-2\alpha}{\beta}$.
Taking arbitrary $0 < \alpha < \frac{1}{2}$ and $C = \sqrt{\frac{\alpha}{4}}$ we see that:
$\mathbb{P}[B] \leq \mathbb{P}[|X-EX| > 2C\sqrt{Var(Z)}]$. Thus we have proved that $p_{2} \geq \delta$
and therefore also $p_{1} \geq \delta$ for some constant $\delta>0$ that does not depend on $\eta$ and $\epsilon$. That was however under an assumption that
$|EX| \leq C\sqrt{Var(Z)}$. 

Now let us assume that $|EX| > C\sqrt{Var(Z)}$. Without loss of generality we can assume that $EX>0$.
In this case we will prove that $\mathbb{P}[|X-EX| < \frac{C}{2}\sqrt{Var(Z)}] < \delta$ for some $0 < \delta <1$
that does not depend on $\eta$ and $\epsilon$. We can proceed as before, using our upper bound on $\mathbb{P}[D_{n} > x]$, to conclude that
$\mathbb{P}[D_{n} < \frac{C}{2}] < \delta$ for some $0 < \delta <1$.
Thus we have proven that
\begin{equation*}
\mathbb{P}_{\kappa=(\kappa_{1},...,\kappa_{n})}[|\sum_{i=1}^{n} \kappa_{i}(y_{i}-h_{i})| > \frac{C}{2} \sqrt{Var(Z)}] > \delta
\end{equation*}
 for some $0 < \delta < 1$ which does not depend on $\eta$ and $\epsilon$.

Denote $T=\frac{C}{2}$. Now we will upper-bound the probability that $\epsilon \sum_{i=1}^{n} \kappa_{i} \geq 
T\sqrt{Var(Z)}$. 
One can easily check that we can assume that: $\epsilon \sum_{i=1}^{n} E\kappa_{i} \leq 
\frac{T}{2}\sqrt{Var(Z)}$ (see: asumptions of the theorem, $Var(Z)$ is large enough).
Now, using standard concentration inequalities such that Chernoff's inequality and again the assumptions of the theorem,
it is easy to verify that the latter probability can be bounded by the negligible function of $R$. That obviously completes the proof.

\textbf{Proof of Theorem 3}

We can proof Theorem 3 by contradiction. 
Note first that the conditions put on the function $\rho(x)$ from Theorem \ref{main_technical_theorem}
are trivially satisfied under our assumptions on the $\kappa$-model.
Let $\{x_{1},...,x_{n}\}$ be a test set.
Without loss of generality we can assume that $\xi$ is as in $\delta$ of Theorem 3 in the main text. 
Denote $h(x_{i})=h_{i}$. Take the groundtruth labels $\{y_{1},...,y_{n}\}$.
Note that an event $\mathcal{E}$ of the form $|\sum_{i=1}^{n} \kappa_{i}(y_{i}-h_{i})| > \sum_{i=1}^{n}\kappa_{i} \epsilon$ is equivalent to saying that the proportion prediction on the bag coming according to the $\kappa$-model was not within error $\epsilon$ from the groundtruth. If we now take the constant $\Delta>0$ for which Theorem \ref{main_technical_theorem} holds and assume by contradiction that the instance label classifier misclassified at least 
$\eta n$ points from the test set for 
$\eta = \frac{\Delta \epsilon^{2}(\sum_{i=1}^{n}E\kappa_{i})^{2}}{n \min_{i=1,...,n} Var(\kappa_{i})}$
then we conclude that the probability of $\mathcal{E}$ is more than $\xi$. 
Note that $E\kappa_i = p_i$, and $Var(p_i) = p_i(1-p_i)$. Taking $\xi$ to be smaller than $\xi(\Delta)$ from Theorem \ref{main_technical_theorem} completes the proof.

\subsection{Private Learning with Label Proportions}

We provide preliminary discussions on the usefulness of LLP in several settings where privacy guarantees are required. 
We will first define an important notion of differential privacy.
Below we give all necessary definitions involving differential privacy and some of its properties.
One can think about differential privacy as a mechanism that, if applied to a given database-access mechanism, 
guarantees small changes in the output as long as the database does not substantially change.

\vspace{-0.1cm}
\begin{definition}[\cite{nissim}]
A randomized algorithm $\mathcal{M}$ satisfies $\epsilon$-differential-privacy if for all
databases $\mathcal{D}_1$ and $\mathcal{D}_2$  
differing on at most one element, and all $S$ $\in$ $Range(\mathcal{M})$,
\begin{equation}
Pr[\mathcal{M} (\mathcal{D}_{1}) = S] \le \exp (\epsilon) \cdot Pr[\mathcal{M} (\mathcal{D}_{2}) = S].
\end{equation}
The probability is taken over the coin tosses of $\mathcal{M}$ .
\end{definition}

Smaller values of $\epsilon$ mean stronger privacy guarantees.
If $f$ is a function and $\mathcal{M}$ aims to compute a differentially-private
version of $f$, it can achieve it by adding noise with the magnitude 
adjusted to the sensitivity of $f$. 

\begin{definition}[\cite{nissim}]
The global sensitivity $S(f)$  of a function $f$ is the smallest
number $s$ such that for all $\mathcal{D}_{1}$ and $\mathcal{D}_{2}$ which differ on at most one element,
\begin{equation}
| f(\mathcal{D}_{1}) - f(\mathcal{D}_{2})\ | \leq s.
\end{equation}
\end{definition}

Let $Lap(0, \lambda) $ denote the Laplace distribution with mean $0$ and 
standard deviation $\lambda$.

\begin{theorem}[\cite{nissim}] \label{laplaciantheorem}
Let $f$ be a function on databases with range $R^{m}$, where $m$ is the number of rows of databases. 
Then, the mechanism that
outputs $f(\mathcal{D} ) + (Y_1,\ldots, Y_m )$, where $Y_i$ are drawn i.i.d 
from $Lap(0,S(f)/\epsilon)$, satisfies $\epsilon$-differential-privacy.  
\end{theorem}

Differential privacy is preserved under composition, but with an extra loss of privacy for each conducted query.
 
\begin{theorem}[\cite{nissim}] \label{compositiontheorem}
\textbf{(Composition Theorem)} The sequential application of mechanisms $\mathcal{M}_i$, each 
giving $\epsilon_{i}$-differential privacy, satisfies $\sum_i \epsilon_{i}$-differential-privacy.  
\end{theorem}

Differential privacy is one of the strongest notions of privacy that is being applied in many different scenarios, 
also in machine learning.
In this section we show how learning from label proportions can make
some existing differentially-private machine learning algorithms even more secure. The mechanism proposed by us below
may serve as a new paradigm for constructing them.

The overall scheme of constructing differentially-private algorithm usually reduces to two steps. In the first step data
is used to construct some machine learning structure. Then this structure with a partial information obtained from the
data is being published. Publishing just partial information enables to get differentially-private guarantees and very often
means outputting some proportions regarding data. For example, in the random decision tree approach 
(see: \cite{weifan}, \cite{jagannathan}) for every leaf of the tree the perturbated counts of items with particular labels
are being released. Then the proportions of those items determine the classification of the test data point.
It should be mentioned however that in the first phase the algorithm producing those proportions has an access to all the data
and stores it explicitly in the structure under construction.
This phase can take significant amount of time and during this period the structure is usually very vulnerable to different
attacks aiming to compromise privacy. 

Note however that since very often the only information involving labels that needs to be
released is the proportions of items with different labels, we can use the classifier that learns the proportions to construct
the structure in the first phase. We feed the learner with bags of data without releasing the actual mapping from instances
to their labels. The information, quantized in those bags, is then used by the learner to compute good approximations of the
proportions and will be later published in a differentially-private way (usually by adding Laplacian noise).

This scheme is more secure than a standard approach since even the algorithm that constructs differentially-private
structure does not know labels of data points it is receiving -- It only knows proportions. So even though the first phase of the
algorithm is definitely not differentially-private (since the algorithm still receives unperturbated data points), it can be easily
proven that it achieves $k$-anonymity. We should note that in the standard approach the first phase did not give any
privacy guarantees.

At the very end we would like to show quantitatively that differential privacy mechanism is a natural candidate for making private the
algorithms that heavily rely on proportions of counts to classify points. Among those algorithm are those involving 
random decision trees or random projection trees \cite{choro, jagannathan}.

Without loss of generality we will assume that we have $2$-labels setting.
Let $\mathcal{S}$ be a set of objects consisting of $n_{+}$ instances with label $+1$ and $n_{-}$ with label $-1$.
Denote $n=n_{+}+n_{-}$. The differentially-private mechanism perturbates counts: $n_{+},n_{-}$ by adding
Laplacian noise independently to each of them to obtain: $n^{p}_{+}$, $n^{p}_{-}$ and 
$n^{p}=n^{p}_{+} + n^{p}_{-}$. 
Denote by $g(\lambda)$ the Laplacian variable with pdf: $\frac{\lambda}{2}e^{-|x|\lambda}$.
The key observation now is that the sensitivity of each count is only $1$. 
Therefore it suffices to add
a Laplacian $g(\frac{\eta}{k})$, where $k$ is the number of all sets $\mathcal{S}$ under consideration 
(we assume that sets $\mathcal{S}$ are pairwise disjoint) and $\eta$ is a desired differential privacy parameter of the
entire algorithm. Thus we have: $n^{p}_{+} = n_{+} + g_{1}$ and $n^{p}_{-}=n_{-}+g_{2}$, where
each $g_{i}$ has the distribution as $g(\frac{\eta}{k})$.
Let fix a set $\mathcal{S}$. Denote: $X = |\frac{n_{+}+g_{1}}{n+g_{1}+g_{2}}-\frac{n_{+}}{n}|$.
Random variable $X$ measures the difference between the actual proportion and the one output by the
differentially-private mechanism.
We have: $X \leq |\frac{g_{1}n_{-}-n_{+}g_{2}}{n^{2}(1+\frac{g_{1}+g_{2}}{n})}|$.
Now one can easily check that for fixed $\eta, k$, $\delta, \theta>0$ and number of instances $n$ large enough
we have: $\mathbb{P}(X > \theta) \leq \delta$. Thus, by Azuma's inequality, with probability at least
$1-e^{-2k\epsilon^{2}}$ all but at most a fraction $\delta+\epsilon$ of all output proportions are within distance
$\theta$ from their nonperturbated versions. 
This means that in practice most of the proportions will be very close to their nonperturbated versions and explains
why differential privacy is a good scalable tool in this setting. The learner that accurately learns label proportions can
make this setting even more secure. It can output proportions close to original without having a straightforward information
about the mapping between the instances used to construct differentially-private structure and their labels.
Those proportions can be then perturbated and published.

\bibliography{LLP}
\bibliographystyle{plain}
\end{document}